\documentclass[11pt,a4paper]{article}
\usepackage{times,latexsym}
\usepackage{url}
\usepackage[T1]{fontenc}

\usepackage[acceptedWithA]{tacl2018v2}

\usepackage{xspace,mfirstuc,tabulary}
\usepackage{dblfloatfix}

\newif\iftaclinstructions
\taclinstructionsfalse 
\iftaclinstructions
\renewcommand{\confidential}{}
\renewcommand{\anonsubtext}{(No author info supplied here, for consistency with
TACL-submission anonymization requirements)}
\newcommand{\instr}
\fi

\iftaclpubformat 

\else

\fi

\usepackage{amsmath,amssymb}
\usepackage{multirow}
\usepackage{graphicx}
\usepackage{pgfplots}
\usepackage{colortbl}
\usepackage{xcolor}

\title{Enriched Attention for Robust Relation Extraction}

\author{Heike Adel \\
	\hspace{7cm}Bosch Center for Artificial Intelligence, Renningen, Germany \\
	{\tt heike.adel@de.bosch.com} \\\And
	Jannik Str\"{o}tgen \\\\
	{\tt jannik.stroetgen@de.bosch.com} \\}

\date{}

\hypersetup{draft}
\begin{document}
	\maketitle
	\begin{abstract}
The performance of relation extraction models
has increased considerably with the rise of neural networks.
However, a key issue of neural relation extraction is robustness:
the models do not scale well to long sentences
with multiple entities and relations. In this work, we address this problem
with an enriched attention mechanism.
Attention allows the model to focus on parts of the
input sentence that are relevant to relation extraction. 
We propose to enrich the attention function
with features modeling knowledge about the 
relation arguments and
the shortest dependency path between them.
Thus, for different relation arguments, the model
can pay attention to different parts of the sentence.
Our model outperforms prior work using comparable setups on two popular
benchmarks,
and our
analysis confirms
that it indeed scales to long sentences with many entities.
\end{abstract}
	\section{Introduction}

Relation extraction is an important task for extracting structured information from unstructured text,
e.g., for knowledge graph population. 
It is typically modeled as a classification task given a sentence and two potential relation arguments, which we call \emph{query entities} in the remainder of the paper.
Current state of the art applies different types of neural networks, such as recurrent neural networks (RNNs) \cite{miwa2016,zhang2017}, convolutional neural networks (CNNs) \cite{zeng2014,zhang2018} or transformer architectures \cite{devlin2019}.
Especially the integration of attention \cite{bahdanau2015} has been shown to improve results \cite{huang2017,zhang2017} as it helps the model to focus on relevant parts of the input.

\begin{figure}
\centering
 \includegraphics[width=.4\textwidth]{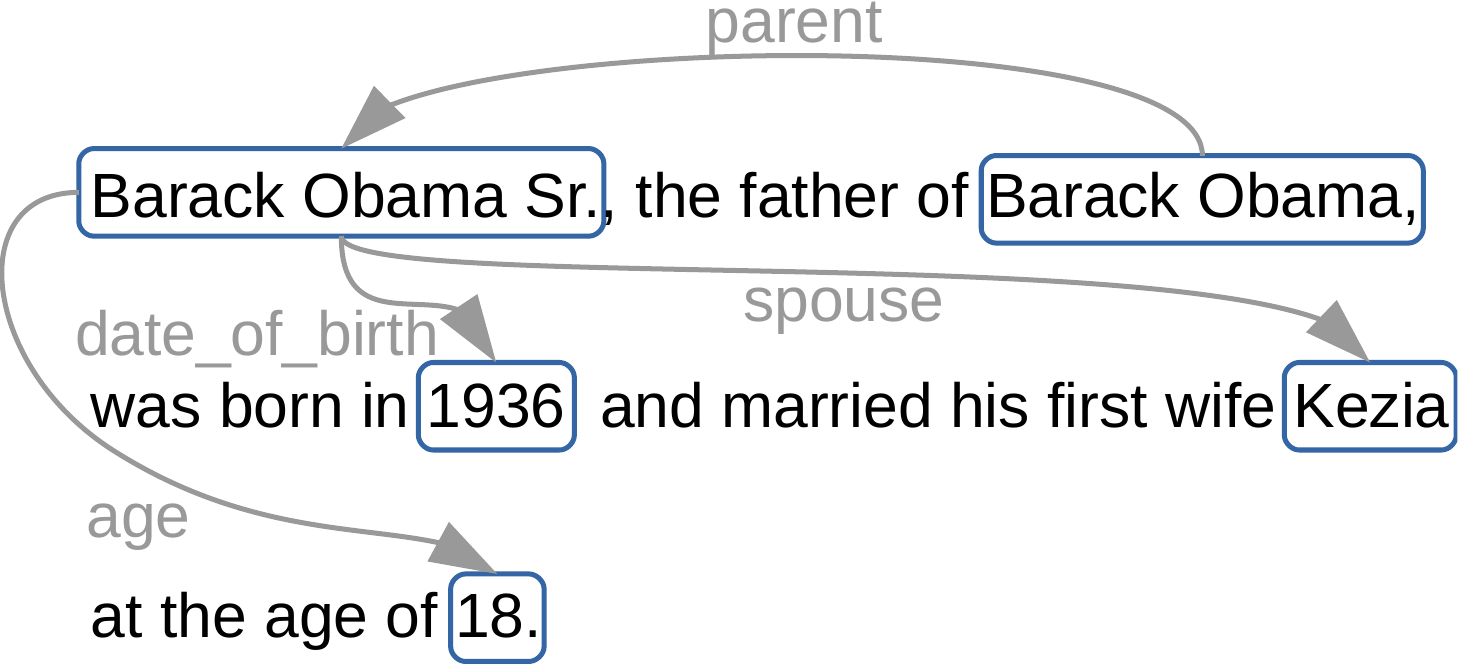}
 \caption{Sample with multiple entities and relations.}
 \label{fig:example}
\end{figure}

For robust relation extraction, however, a model
needs to detect not only key phrases like ``was born in'',
``married'' or ``at the age of'' but also has to determine to which entities they belong.
This is particularly challenging when the context between two 
query entities 
includes other entities, as in the 
sentence
in Figure \ref{fig:example}: 
The context between ``Barack Obama Sr.''
and ``1936'', for example, includes the entity ``Barack Obama''.
If an automatic extraction model only learns relation-specific key words
or key phrases (as max pooling or attention layers do), it will struggle with such long sentences
containing diverse information. From the sample sentence, a key word-based model
might extract, for example, the 
incorrect 
relation
\texttt{born\_in(Barack Obama, 1936)} instead of the correct one 
\texttt{born\_in(Barack Obama Sr., 1936)}.

In this paper, 
we study the possibility to enrich the attention layer of a state-of-the-art model with additional information (features)
to increase the robustness of attention-based relation extraction models.
In particular, we address the task with a long-short term memory network architecture and investigate two different attention functions: \emph{Additive attention} \cite{bahdanau2015} as also used by \newcite{zhang2017} and \emph{dot-product attention} which gets more and more popular nowadays due to transformer networks \cite{vaswani2017}. 
Previous work on relation extraction \cite[i.a.,][]{fundel2006,mintz2009,rink2010,miwa2014} showed the benefit of features derived from (i) the shortest path between query entities in a dependency parse tree and (ii) information about the types of the query entities.
Our approach builds upon these observations. In particular, we investigate the following features for the attention function: 
for \emph{dependency-enriched attention}, a representation of the \emph{shortest path} between the two query entities,	as well as several token-level features, such as the \emph{distance} of each token to the query entities in the dependency parse tree;
for \emph{entity-enriched attention}, 
embeddings representing the \emph{query entities}, 
and embeddings representing their \emph{types}.

Our model outperforms prior work using comparable setups on TACRED and achieves state-of-the-art results on ACE 2005,
two popular benchmark datasets for relation extraction.
Our analysis shows that the additional features 
help the network to pay more attention to words which are relevant to the two query entities but not to other entities in the sentence, alleviating the challenges shown in Figure \ref{fig:example}. 
Thus, our work increases the scalability of relation extraction models to longer sentences with mentions of multiple entities and relations.

In summary, our main contributions are:

\begin{itemize}
	\item We propose to enrich the attention mechanism with additional features.
	This way, the model can attend to the parts of the sentence which are
	relevant to the
	query entities and ignore other parts which might include relation information about other entities.

	\item Our approach is broadly applicable. Although we train long short-term memory networks in this work, our 
	enriched attention mechanism is complementary to the underlying model and
	can thus be included in any attention-based neural network. Similarly, the concept of enriched attention is general and
	can be used for other tasks as well.

	\item We outperform prior work with comparable setups on two popular benchmarks and perform an in-depth analysis demonstrating that our enriched attention methodology indeed increases the robustness of the model in the case of long sentences and multiple entities per sentence.
\end{itemize}
	\section{Related Work}
\textbf{Neural relation extraction.}
Related work builds upon different input representations:
the whole sentence \cite{zhang2017},
a combination of three contexts \cite{adel2019},
the textual context between the query entities \cite{zeng2014},
entity graphs \cite{christopoulou2019}, span graphs \cite{luan2019},
the dependency tree of the sentence \cite{guo2019} 
or the shortest dependency path between the query entities \cite{xu2015,toutanova2015,xu2015,cai2016,huang2017,zhang2018}.
\newcite{pouran2020}
combine information from the dependency tree with the textual context.
In contrast, we use the 
whole sentence as input and utilize information from the dependency tree 
as features for the attention function.
Another 
feature 
known to improve 
relation extraction is type information of query entities
\cite{mintz2009,yao2010,ling2012,DelCorroEtAl2015,lin2020}. We propose to
integrate type information into the attention layer as further features.

Previous work 
uses different architectures, such as RNNs \cite{miwa2016,zhang2017}, CNNs \cite{zeng2014,huang2017,zhang2018}, graph convolutional neural networks (GCNs) \cite{guo2019} or transformer networks \cite{alt2019,soares2019}. 
Note that the latter approach typically requires extensive pre-training.
Attention 
for relation extraction is not only popular in the context of transformers \cite{vaswani2017,devlin2019} but also 
with other models, such as CNNs \cite{huang2017}, GCNs \cite{guo2019} or RNNs \cite{zhang2017}.
We use the latter approach as a baseline model.

\textbf{Extending the attention function.}
Enriching attention for natural language processing was proposed by, i.a., \newcite{li2019} via incorporating relation indicating keywords into the attention function, by \newcite{adel2017} via exploiting external information from gazetteers, by \newcite{zhong2019} via integrating common sense knowledge into transformer models and by \newcite{zhang2017} via position-aware attention.
In contrast, we propose to integrate a variety of
dependency tree- and entity type-based features
into 
attention 
to increase the robustness of the model
and to be able to cope with sentences consisting of multiple entity pairs.
Our approach is purely data-driven and does not require knowledge about the target relations. This is in contrast to the approach by \newcite{li2019}.
Moreover, we also compare two different attention functions and show that our features can be integrated into both of them.
	\section{Model}
We describe the basic model in Section~\ref{subsec:baselineModel}
and explain the enriched attention layer as well as the different features in
Section~\ref{subsec:enrichedattention}.

\subsection{Baseline Model for Relation Extraction}
\label{subsec:baselineModel}
We build upon the model by \newcite{zhang2017},
one of the state-of-the-art 
approaches for relation classification.
The model consists of four main components: The input layer, 
two stacked LSTM layers, an attention layer, and the output layer.
To enable a direct comparison to prior work, we take input, LSTM, and output layers from \newcite{zhang2017} and only replace their
attention layer by our enriched attention layer 
(see Section \ref{subsec:enrichedattention}).\footnote{We also experimented with using a transformer network instead of an LSTM network but achieved lower results on the development set. We assume the reason is that we trained the transformer network from scratch without any pre-training.}

\textbf{Input layer.} Each token is represented by a concatenation
of its word embedding, its part-of-speech embedding, and its named-entity-tag embedding.
As word embeddings, we use pre-trained 300-dimensional GloVe embeddings\footnote{\url{http://nlp.stanford.edu/data/glove.840B.300d.zip}} \cite{pennington2014} as well as a concatenation of GloVe and ELMo embeddings\footnote{\url{https://github.com/allenai/allennlp/blob/master/allennlp/modules/elmo.py}} \cite{peters2018}.
Part-of-speech and named-entity-tag embeddings are randomly initialized.
All embeddings are fine-tuned during training. 

\textbf{LSTM layers.} The token representations are fed into two stacked
unidirectional LSTM layers
to obtain context-aware representations.
To compute a representation for the whole sentence, the 
hidden states of the last LSTM layer are combined using a weighted sum with 
attention weights.

\textbf{Output layer.} Finally, the sentence representation is fed into a linear layer
which maps it to a vector of the number of output classes.
A softmax activation function is used to obtain a probability distribution
over the output classes.

\begin{figure*}
\centering
 \includegraphics[width=.85\textwidth]{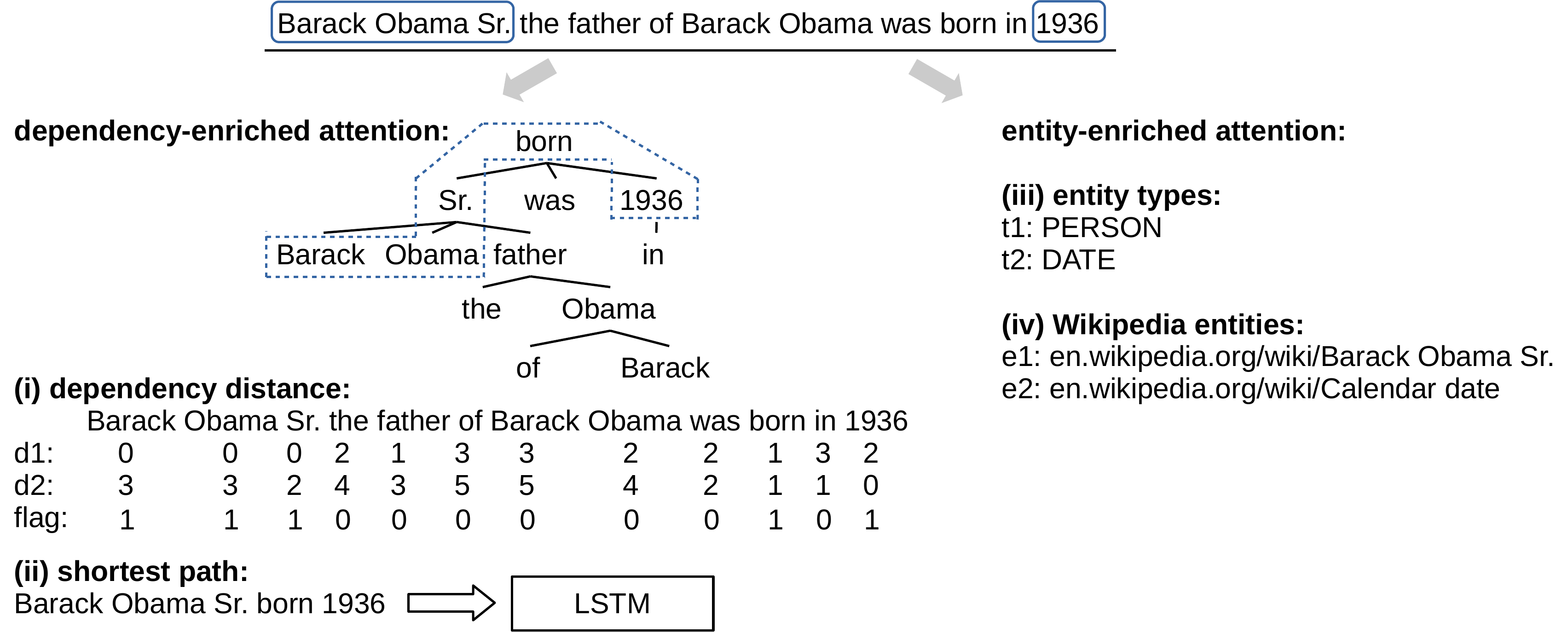}
 \caption{Features for enriched attention for a sample sentence. The query entities ``Barack Obama Sr.'' and ``1936''
 are marked with boxes, the shortest path between them in the dependency parse tree is marked with a dashed box.}
 \label{fig:features}
\end{figure*}

\subsection{Enriched Attention Layer}
\label{subsec:enrichedattention}
Models for relation extraction often do not scale well to long sentences with mentions of multiple entities and relations.
This is especially harmful when relation extraction is used as a component for a downstream task, such as knowledge base population.
Previous work, such as \newcite{huang2017} or \newcite{adel2019} found this to be one of the most frequent mistakes in the relation extraction part of slot filling systems.
Therefore, we study the integration of various additional signals into the attention layer of relation extraction models, such as information from the dependency parse tree and information about the query entities. 
We now describe our proposed enriched attention layer
and its features in detail. The features are illustrated in Figure \ref{fig:features}.

\subsubsection{Attention Functions}

We distinguish between two different levels for the integration of features into 
the attention layer: \emph{local} and \emph{global}.
Local features are token-specific, i.e., for
calculating the attention weights of hidden state $i$, we use different features
than for calculating the attention weights of hidden state $j$ ($i \neq j$).
Global features are sentence-specific, i.e., they are the same for
calculating the attention weights of all hidden states for a particular 
input sentence.

Given local features $l_i \in \mathbb{R}^{L}, 1 \le i \le n$
with $n$ being the sentence length,
and global features $g \in \mathbb{R}^G$, 
we compare their integration into two different attention functions: additive attention using an additional hidden layer as proposed by \newcite{bahdanau2015} (called $\oplus$-att in the remainder of the paper) and dot-product attention as used in transformer networks \cite{vaswani2017} (called $\odot$-att).

In both cases, attention weights $\alpha_i$ for each token $i$ are calculated by normalizing attention scores $e_i$ (see Equations \ref{eq:main} and \ref{eq:self-att}) using the softmax function:

\begin{equation}
\alpha_i = \frac{\exp(e_i)}{\sum_{j = 1}^n\exp(e_j)}
\label{eq:alpha}
\end{equation}

\paragraph{$\oplus$-att.}
The attention score $e_i$ for token $i$ is calculated as follows: 
\begin{align}
e_i  = & \;
v^\top \tanh \big( W_h h_i + W_q q + W_s p_i^s + W_o p_i^o \notag
\\ & \;\;\;\;\;\;\;\;\;\;\;\; + W_l l_i + W_g g \big)
\label{eq:main}
\end{align}
with $v \in \mathbb{R}^A$, $W_h \in \mathbb{R}^{A \times H}$, 
$W_q \in \mathbb{R}^{A \times H}$,
$W_s, W_o \in \mathbb{R}^{A \times P}$,
$W_l \in \mathbb{R}^{A \times L}$,
$W_g \in \mathbb{R}^{A \times G}$ being the trainable parameters of the attention layer and
$A$ being its hyperparameter.
The hyperparameters $L$ and $G$ are the dimension of the local features ($l_i \in \mathbb{R}^L$) and
the dimension of the global features ($g \in \mathbb{R}^G$), respectively.
The LSTM hidden state size from the previous layer is denoted by $H$:
$h_i, q \in \mathbb{R}^H$ with $h_i$ being
the hidden state corresponding to token $i$.
Following \newcite{zhang2017}, we also integrate $q$, the last hidden state of the sentence, and position features $p_i^s, p_i^o \in \mathbb{R}^P$ where $p_i^s$ encodes the distance of token $i$ to the first query entity and $p_i^o$ encodes the distance of token $i$ to the second query entity.
The dimension of the position features is called $P$.

\paragraph{$\odot$-att.}
With dot-product attention, the attention score $e_i$ for token $i$ is calculated as follows: 
\begin{align}
e_i = & \frac{q_i \cdot k_i + l_i \cdot l_i + g \cdot g}{\sqrt{d}}\;
\label{eq:self-att}
\end{align}
with $q_i$, $k_i$ being the query and key vectors, computed as in \newcite{vaswani2017} with learned weight matrices $W^Q$, $W^K$ based on the LSTM hidden states: $q_i = W^Qh_i$, $k_i = W^Kh_i$. 
Similarly, a value vector $v_i = W^Vh_i$ is computed and later multiplied with the attention weights $\alpha_i$ (cf., Equation \ref{eq:alpha}). 
Following \newcite{vaswani2017}, we scale the attention score with the square root of $d$, the dimension of the key and query vectors.

\subsubsection{Dependency-Enriched Attention}
\label{sec:dependencyfeatures}
Previous work, especially work on non-neural models \cite[i.a.,][]{bunescu2005,mintz2009,rink2010,miwa2014} has shown that features based on the results of dependency parsing are beneficial for relation extraction. The intuition is that the shortest path between two entities in a dependency parse tree contains relevant information about the relation between them and less noise than the text sequence between them. This is shown in Figure \ref{fig:features} as well: While the text sequence between the query entities ``Barack Obama Sr.'' and ``1936'' also contains information about ``Barack Obama'', the shortest path between the query entities only consists of the word ``born'' which is relevant for determining the correct relation.

Some prior work on neural relation extraction uses the shortest dependency
path between the two query entities directly as the input for the neural network
\cite[e.g.,][]{toutanova2015,huang2017,zhang2018}.
We argue that there is a risk of losing information, 
for example, when
the automatically extracted dependency parse tree is wrong. 
Therefore, we do not replace the original input sentence
with a potentially noisy dependency path.
Instead, we propose to use information from the dependency parse tree as an additional signal for the attention layer.
As a result, the model can learn to pay more attention to the tokens on the shortest dependency path between the two query entities but it can also decide to ignore them if they are noisy or not helpful.

Given the dependency parse tree, we explore the following features for attention which are shown in the left part of Figure \ref{fig:features} for a sample sentence and two query entities.

\textbf{(i) Dependency distance:}
For each token, we calculate its \emph{distance} to the two query entities in the dependency parse tree.
We represent the distances with embeddings $d^{e_1} \in \mathbb{R}^D$ and $d^{e_2} \in \mathbb{R}^D$ that are randomly initialized.
The embedding dimension $D$ is tuned on the development (dev) set. 
Note that the distance embeddings are not shown in Figure \ref{fig:features} for 
the sake of readability.

Furthermore, we use a \emph{flag} $f \in \mathbb{R}^1$ (binary, single-valued feature) to indicate
whether a token is on the shortest dependency path between the two query entities.

Both the distances and the flag are token-dependent, i.e., local features.
To integrate them in Equation \ref{eq:main}, we concatenate them to one vector per token 
$l_i := [d_{i}^{e_1}; d_{i}^{e_2}; f_i] \in \mathbb{R}^{2D+1}$ with $[;]$ denoting vector concatenation.

\textbf{(ii) Shortest path:}
Given a sentence and two query entities, we extract the \emph{shortest path}
between the two entities
from the dependency parse tree (lower left part of Figure \ref{fig:features}). 
To embed the shortest dependency path in vector space, we 
first represent the tokens on the path
by their word embeddings and then feed them into an LSTM model
with hidden state size $S$.
The final representation $s \in \mathbb{R}^{S}$ of the shortest dependency path is the 
last hidden state of the LSTM.

Note that we use the same word
embeddings as in the input layer of the network but the parameters of
the LSTM model over the path are additional parameters, i.e., not shared 
with the sentence-level LSTM model
from Section \ref{subsec:baselineModel}.
The reason is two-fold:
First, the linguistic structure of the shortest dependency
path is inherently different to the structure of a sentence. Thus, trying to learn both
structures using the same parameters could harm both sub-tasks. Second, we assume that
it would put too much weight on a particular feature if we used the same 
distributions for the input representation and that feature.
Thus, using independent weights for the second LSTM
allows the attention layer to focus not only on the shortest dependency path but also on other
signals.

The path is independent of the particular input token, i.e., a global feature. 
To integrate it in Equation \ref{eq:main} or \ref{eq:self-att}, we set $g := s$.

\subsubsection{Entity-Enriched Attention}
\label{sec:entityfeatures}
Many studies on relation extraction show the positive impact of 
exploiting information about the types of the query entities
\cite{mintz2009,ling2012}.
This can be explained by the intuition that knowledge about the types can reduce the search space of possible relations \cite{roth2004,yao2010,ren2017}.
In the example of Figure \ref{fig:features}, knowing that the second query entity is a date can help the model exclude relation types like city\_of\_birth which would require a query entity of type location.
Here, we explore the applicability of entity type features in the context of attention.
Since specific relation key phrases like ``was born in'' are only applicable
to persons and locations but not to, e.g., organizations, the model
can learn to ignore those phrases for organization entities but
attend to them for persons and locations.
The query entities are sentence-specific. Thus, the resulting features 
(see right part of Figure \ref{fig:features})
are global features. 

\textbf{(iii) Entity types:}
We use embeddings $t^{1}, t^{2} \in \mathbb{R}^{T}$ to 
represent the types of the two entities.
Thus, $g$ from Equations \ref{eq:main} and \ref{eq:self-att} becomes a concatenation of the two embeddings:
$g := [t^1; t^2]$.
The embeddings are randomly initialized and learned during training.
Their dimensionality $T$ is tuned on dev.

\textbf{(iv) Wikipedia entities:} 
The last feature we explore is a representation
of the query entities themselves.
To avoid overfitting to the entities of the training set,
we do not learn entity representations from scratch but 
use pre-trained embeddings for Wikipedia entities \cite{yamada2017}.
Section \ref{sec:preprocessing} describes how we map 
the query entities to Wikipedia entities and handle unknown entities.

When integrating entity embeddings into Equation \ref{eq:main} or \ref{eq:self-att},
the global feature $g$ becomes $[e^1; e^2]$ with $e^1, e^2 \in \mathbb{R}^{300}$
being the entity embeddings.
	\section{Experiments and Analysis}
In this section, we describe
the datasets and our experiments,
and presents our results and analysis.

\begin{table}
	\centering
	\footnotesize
	\begin{tabular}{l|r|r}
		& TACRED & ACE \\
		\hline
		\# instances & 106,264 & 87,805 \\
		\# relations & 41+1 & 12+1\\
		\# no-relation instances & 79.5\% & 92.0\%\\
		avg sentence length & 36.4 & 33.7 \\
	\end{tabular}
	\caption{Dataset statistics.}
	\label{tab:data}
\end{table}

\begin{table}
	\footnotesize
	\begin{tabular}{p{.01cm}l|l}
		\multirow{6}{*}{\rotatebox{90}{General model}}
		& Glove word emb dim & 300\\
		& NER tag emb dim & 30\\
		& POS tag emb dim & 30\\
		& Position emb dim & 10\\
		& LSTM hidden units & 200\\
		& \# LSTM layers & 2 (unidirectional)\\
		\hline
		\multirow{7}{*}{\rotatebox{90}{Attention}}
		& Attention layer hidden units & 100\\
		& $\odot$-attention \# heads & 1\\
		& Distance emb dim & 10\\
		& SP-LSTM hidden units & 200\\
		& \# SP-LSTM layers & 1 (unidirectional)\\
		& Entity type emb dim & 30\\
		& Wikipedia entities emb dim & 300\\
		\hline
		\multirow{8}{*}{\rotatebox{90}{Training}}
		& Optimizer & SGD\\
		& Batch size & 50\\
		& Initial learning rate & 1.0\\
		& Learning rate decay & 0.9\\
		& Gradient clipping & 5.0\\
		& Dropout rate & 0.5\\
		& Word dropout rate & 0.04\\
		& Max number of epochs & 30 (TACRED), 60 (ACE)\\
	\end{tabular}
	\caption{Training and model hyperparameters. SP: shortest dependency path, emb dim: embedding dimension.}
	\label{tab:hyperparams}
\end{table}

\subsection{Data}
\label{sec:data}
We evaluate our model on 
popular benchmarks 
for
relation extraction: TACRED and ACE 2005.\footnote{LDC2018T24 and LDC2006T06.}

\textbf{TACRED.}
The large-scale TAC Relation Extraction Dataset \cite{zhang2017}
has been annotated via crowd-sourcing.
It is a classification dataset in which only one entity pair per sentence is annotated with a relation.
In total, it covers 41 relations between persons, organizations,
geo-political entities, dates, numbers and nominal phrases like
cause of death. 
Entity pairs without relations are labeled with \texttt{no\_relation}.

\textbf{ACE.}
ACE 2005 \cite{walker2006} is a manually labeled dataset with six relations between
persons, organizations, geo-political entities, locations, facilities, weapons and vehicles.
For each sentence, all occurring entities and relations are annotated.
Unlike TACRED, it does not provide a split into training, development and test set.
To compare with prior work, we use the split by \newcite{miwa2016} and \newcite{li2014}.
Following those previous work,
we consider all entity pairs
in a sentence as classification instances and introduce inverse relations (as independent class labels)
to model the direction of relations. Thus, we consider twelve relations in total.
All entity pairs not occurring in the 
manual annotations are labeled with
\texttt{no\_relation}.

\textbf{Statistics.}
While 
TACRED
consists of more relation 
classes, 
the main challenge of the ACE dataset is that
relations between all entity pairs have to be extracted for each sentence.
This results in very similar classification instances the model needs 
to distinguish. 
It also leads to a considerably higher amount of instances
of the \texttt{no\_relation} class 
and, thus, a class imbalance challenge.
The sentence lengths are rather long in both datasets.
See Table \ref{tab:data} for more statistics.

\begin{table}
	\footnotesize
	\centering
	\begin{tabular}{p{.03cm}l|c|c|c|c}
		& 	& dev & \multicolumn{3}{c}{test}\\
		& 	Model & $F_1$ & $P$ & $R$ & $F_1$\\
		\hline
		\multirow{10}{*}{\rotatebox{90}{single models}}
		&	\cite{zhang2017} & 66.0 & 65.7 & 64.5 & 65.1\\
		&	\cite{zhang2018} & 67.4 & 69.9 & 63.3 & 66.4\\
		&	\cite{alt2019} & 68.0 & 70.1 & 65.0 & 67.4\\
		&	\cite{li2019} & - & 67.1 & 68.4 & 67.8\\
		&	\cite{guo2019} & - & 71.8 & 66.4 & 69.0\\
		&	\cite{soares2019} & 70.1 & - & - & 70.1\\
		\cline{2-6}
		&	enriched $\oplus$-att & 66.8 & 65.3 & \textbf{68.9} & \textbf{67.1}\\ 
		&	+ ELMo & 66.6 & 63.9 & 65.7 & 64.8\\ 
		&	enriched $\odot$-att & 66.5 & 67.0 & 65.3 & 66.1\\ 
		&	+ ELMo & \textbf{67.3} & \textbf{69.3} & 63.2 & 66.1 \\ 
		\hline 
		\multirow{6}{*}{\rotatebox{90}{ensembles}}
		&	\cite{zhang2017} & - & 70.1 & 64.6 & 67.2\\
		&	\cite{zhang2018} & - & 71.3 & 65.4 & 68.2\\
		\cline{2-6}
		&	enriched $\oplus$-att & 68.1 & 67.1 & 68.1 & 67.6\\
		&	+ ELMo & 68.6 & 66.2 & \textbf{68.8} & 67.5\\
		&	enriched $\odot$-att & 68.2 & \textbf{70.7} & 65.2 & 67.8\\
		&	+ ELMo & \textbf{69.3} & 70.4 & 65.4 & \textbf{68.3}\\ 
	\end{tabular}
	\caption{Results on TACRED data. Bold highlights our best configurations.}
	\label{tab:resultsTACRED}
\end{table}

\subsection{Pre-processing for Enriched Attention}
\label{sec:preprocessing}
Our features for enriched attention build
on the following three pre-processing steps.

\begin{table}
	\footnotesize
	\centering
	\begin{tabular}{l|c|c|c}
		Model & $P$ & $R$ & $F_1$\\
		\hline
		(i) dependency distance & 67.3 & 64.8 & 66.0\\
		(ii) shortest path & 67.2 & 64.7 & 65.9\\
		(i) + (ii) & 66.9 & 65.4 & 66.1\\
		(iii) entity types & \textbf{68.0} & 64.4 & 66.2\\  
		(iv) Wiki embeddings & 66.0 & 66.1 & 66.0 \\
		(iii) + (iv)  & 65.0 & \textbf{67.5} & 66.2\\
		(i) + (ii) + (iii) + (iv) & 66.2 & 66.9 & \textbf{66.5}\\
	\end{tabular}
	\caption{Variations of enriched $\odot$-attention on TACRED development set.} 
	\label{tab:ablationTACRED}
\end{table}

\begin{table}
	\footnotesize
	\centering
	\begin{tabular}{p{.01cm}l|c|c|c|c}
		& & dev & \multicolumn{3}{c}{test}\\
		& Model & $F_1$ & $P$ & $R$ & $F_1$ \\
		\hline
		\multirow{8}{*}{\rotatebox{90}{single models}} &
		\cite{li2014} & - & 68.9 & 41.9 & 52.1\\
		& (Miwa et al., 2016)* & - & 70.1 & 61.2 & 65.3\\
		& \cite{sun2019} & - & 68.7 & 65.4 & 67.0\\
		& \cite{ye2019}** & - & 66.5 & 71.8 & 68.9\\
		\cline{2-6}
		& enriched $\oplus$-att & 61.0 & 65.9 & 55.5 & 60.3\\
		& + ELMo & \textbf{72.8} & 73.3 & \textbf{75.3} & 74.3\\ 
		& enriched $\odot$-att & 67.9 & 68.6 & 68.0 & 68.3\\ 
		& + ELMo & 72.7 & \textbf{76.9} & 73.3 & \textbf{75.1}\\ 
		\hline 
		\multirow{4}{*}{\rotatebox{90}{ensembles}}
		& enriched $\oplus$-att & 64.2 & 71.4 & 60.1 & 65.3\\
		& + ELMo & \textbf{75.0} & 75.5 & \textbf{74.7} & 75.1\\
		& enriched $\odot$-att & 69.2 & 71.3 & 70.2 & 70.7\\
		& + ELMo & 74.4 & \textbf{76.9} & 74.5 & \textbf{75.7}\\
	\end{tabular}
	\caption{Results on ACE 2005 test data with gold entities. 
		Bold highlights our best configurations.
		* result of \newcite{miwa2016} for relation extraction using gold entities is reported in \cite{christopoulou2019}. ** does not consider directionality of relation. }
	\label{tab:resultsACE}
\end{table}

\begin{table*}
	\footnotesize
	\centering
	\begin{tabular}{l|c|c|c|c|c|c|c|c}
		& \multicolumn{4}{c|}{Entities} & \multicolumn{4}{c}{Relations}\\
		& dev & \multicolumn{3}{c|}{test} & dev & \multicolumn{3}{c}{test}\\
		Model & $F_1$ & $P$ & $R$ & $F_1$ & $F_1$ & $P$ & $R$ & $F_1$ \\
		\hline
		\cite{li2014} & - & 85.2 & 76.9 & 80.8 & - & \textbf{65.4} & 39.8 & 49.5\\
		\cite{miwa2016} & 81.8 & 82.9 & 83.9 & 83.4 & 51.8 & 57.2 & 54.0 & 55.6\\
		\cite{sun2018} & - & 83.9 & 83.2 & 83.6 & - & 64.9 & 55.1 & 59.6\\
		\cite{li2019qa} & - & 84.7 & 84.9 & 84.8 & - & 64.8 & 56.2 & 60.2\\
		\hline
		enriched $\odot$-att & \textbf{86.9} & \textbf{89.1} & \textbf{89.0} & \textbf{89.0} & 56.6 & 57.2 & 59.3 & 58.2\\
		enriched $\odot$-att + ELMo & \textbf{86.9} & \textbf{89.1} & \textbf{89.0} & \textbf{89.0} & \textbf{59.6} & 59.6 & \textbf{64.8} & \textbf{62.1}\\
	\end{tabular}
	\caption{Results on ACE 2005 test data for joint entity and relation extraction. Our scores are evaluated in a pipeline setting, using the relation extraction model with enriched attention based on the entities found with a bi-directional LSTM+CRF model.}
	\label{tab:resultsJointACE}
\end{table*}

\textbf{Dependency parsing.}
TACRED already provides the results of 
various pre-processing steps with Stanford CoreNLP \cite{manning2014} including automatic dependency parsing.
We preprocess the ACE 2005 accordingly.

\textbf{Entity typing.}
Both datasets provide
entity type information.
Examples 
are \texttt{PERSON},
\texttt{DATE}, \texttt{NATIONALITY} and \texttt{TITLE} in TACRED,
and
\texttt{PER} (person),
\texttt{FAC} (facility) and \texttt{VEH} (vehicle)
in ACE 2005.
For a dataset without given entity types, types can be obtained
automatically, for example using CoreNLP or 
fine-grained typing methods \cite{ling2012,DelCorroEtAl2015,yaghoobzadeh2018}.

\textbf{Entity linking.}
To map the entities of our datasets to Wikipedia entities, 
we apply the AIDA entity linking system 
\cite{HoffartEtAl2011}.
To model out-of-knowledge-base entities with a 
meaningful representation in the same space,
we 
use the embedding for the Wikipedia page about the type
of the entity. The rationale behind that is that such embeddings represent a prototypical 
description of the respective 
entity types.
An example 
is 
shown
in Figure~\ref{fig:features}: Since the 
mention ``1936'' 
is not 
linked to a Wikipedia entity,\footnote{Although
there is a Wikipedia article for the year ``1936'', most state-of-the-art
entity linking systems do not consider numerical values / temporal expressions for linking.} 
the model
uses the embedding
of the type \texttt{DATE}, i.e., of the page
\url{en.wikipedia.org/wiki/calendar\_date}.
Mappings from types to Wikipedia pages are provided in the appendix.

\subsection{Training Details}
We train our model with mini-batch stochastic gradient descent, using a batch size
of 50 and an initial learning rate
of 1.0.
We 
decrease the learning rate exponentially with a factor of 0.9 based
on the performance of the model on the development set.
For regularization, we apply dropout \cite{srivastava2014} with a probability of 0.5
as well as word dropout which randomly replaces tokens with the unknown token symbol with a probability of 0.04.
To cope with exploding gradients, we clip the gradients at a threshold of 5.0.

The GloVe word embeddings and the ELMo word embeddings have a dimensionality of 300 and 1024, respectively.\footnote{We take the average of forward and backward embeddings from ELMo as word embeddings.}
The dimensions of both the named-entity-tag embeddings and part-of-speech-tag embeddings are 30.
Each sentence is processed
with a 2-layer unidirectional LSTM model with 200 hidden units 
per layer.
The hyperparameters of enriched attention are tuned with grid search on the TACRED development set
and used on both datasets.
A complete list with all hyperparameters is provided in Table \ref{tab:hyperparams}.

\begin{figure*}
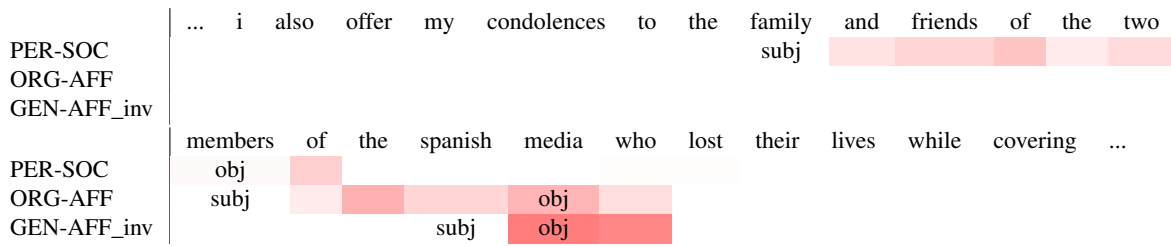

	\footnotesize
	\begin{tabular}{l|cccccccccccccc}
		& ... & i & also & offer & my & condolences & to & the & family & and & friends & of & the & two \\
		PER-SOC &  \cellcolor{red!0} & \cellcolor{red!0}&  \cellcolor{red!0} & \cellcolor{red!0} & \cellcolor{red!0} & \cellcolor{red!0} & \cellcolor{red!0}&  \cellcolor{red!0} & \cellcolor{red!0}subj &\cellcolor{red!11} &\cellcolor{red!16} &\cellcolor{red!23}& \cellcolor{red!8} &\cellcolor{red!14}\\
		ORG-AFF &  \cellcolor{red!0} & \cellcolor{red!0}&  \cellcolor{red!0}& \cellcolor{red!0} &\cellcolor{red!0} &\cellcolor{red!0} & \cellcolor{red!0} & \cellcolor{red!0} & \cellcolor{red!0} &\cellcolor{red!0} &\cellcolor{red!0} &\cellcolor{red!0} &\cellcolor{red!0}& \cellcolor{red!0}\\
		GEN-AFF\_inv &\cellcolor{red!0} &\cellcolor{red!0} &\cellcolor{red!0} &\cellcolor{red!0}& \cellcolor{red!0} &\cellcolor{red!0}& \cellcolor{red!0}& \cellcolor{red!0} & \cellcolor{red!0} &\cellcolor{red!0} &\cellcolor{red!0} &\cellcolor{red!0} &\cellcolor{red!0}& \cellcolor{red!0}\\
	\end{tabular}
	
	\begin{tabular}{l|cccccccccccccc}
		&  members & of & the & spanish & media & who & lost & their & lives & while & covering & ... \\
		PER-SOC  &\cellcolor{red!2}obj &\cellcolor{red!19} &\cellcolor{red!0} &\cellcolor{red!0} & \cellcolor{red!0} & \cellcolor{red!1} &\cellcolor{red!1} &\cellcolor{red!0}& \cellcolor{red!0} &\cellcolor{red!0}& \cellcolor{red!0} \\
		ORG-AFF  &\cellcolor{red!0}subj &\cellcolor{red!8} &\cellcolor{red!31} &\cellcolor{red!16} & \cellcolor{red!29}obj& \cellcolor{red!13} &\cellcolor{red!0} &\cellcolor{red!0} &\cellcolor{red!0} &\cellcolor{red!0} &\cellcolor{red!0} \\
		GEN-AFF\_inv  &\cellcolor{red!0}& \cellcolor{red!0} &\cellcolor{red!0} &\cellcolor{red!0}subj & \cellcolor{red!51}obj &\cellcolor{red!47} &\cellcolor{red!0} &\cellcolor{red!0} &\cellcolor{red!0} &\cellcolor{red!0} &\cellcolor{red!0}
	\end{tabular}
	\caption{Attention weights for sentence with different query entities from ACE development set. First column shows gold relation labels (not known to the model).}
	\label{tab:analysis_weights}
\end{figure*}

\subsection{Results}
Following \newcite{zhang2017}, we report the test score that corresponds
to the medium dev score out of five runs for each model variation.
Following prior work on TACRED, we also report results of model ensembles.
In particular, we train the same model configuration five times with different seeds for weight initialization.
Afterwards, we merge their results using majority vote.

\textbf{Results on TACRED.}
Table \ref{tab:resultsTACRED}
provides the results of our model on the TACRED dataset in comparison to state of the art. 
Our model is best comparable to the model by \newcite{zhang2017} as it uses the same basic model components and the $\oplus$-att attention function is a direct extension of their position-aware attention. The results show that this extension increases the performance by 2 F1 points.
Interestingly, adding ELMo embeddings does not further increase performance.
Furthermore, our model performs better than the model by \newcite{zhang2018} who use dependency tree information on the input level while we use the original text sequence as input and incorporate dependency tree information into the attention function.
Our results do not reach the performance of the current state of the art by \newcite{soares2019} since we train our model from scratch on only the TACRED data while \newcite{soares2019} fine-tune a BERT model. Thus, their model is both much larger and more expensive to train and utilizes information from a large pre-training corpus.

In Table \ref{tab:ablationTACRED}, we show an ablation study for different configurations of enriched $\odot$-attention. 
In particular, we investigate the impact of using different subsets of features for the attention function.
The combination of all features leads to the best performance, indicating that all of them are useful for relation extraction.

\textbf{Results on ACE.}
Tables \ref{tab:resultsACE} and \ref{tab:resultsJointACE} provide the results of our model on the ACE dataset in comparison to state of the art.
We focus on systems that use the same or similar dataset split as we do and are, thus, directly comparable. 
There is another line of work on the ACE dataset which splits the dataset with respect to domains in order to investigate cross-domain generalization of models \cite[e.g.,][]{nguyen2016,pouran2020}.
Since this introduces an incomparable experimental setup, we do not compare to those works here.

While Table \ref{tab:resultsACE} shows results for relation extraction with gold entities, Table \ref{tab:resultsJointACE} shows results for joint entity and relation extraction.
For the latter, we use a pipeline setting with a standard state-of-the-art named entity recognition (NER) model, consisting of a bi-directional LSTM layer followed by a conditional-random-field output layer \cite[e.g.,][]{lample2016}.
The input tokens for the NER model are represented by a concatenation of Flair \cite{akbik2018}, byte-pair-encoding \cite{heinzerling2018}, GloVe \cite{pennington2014} and XLM\_Roberta embeddings \cite{conneau2020}.
On ACE, enriched $\odot$-attention outperforms enriched $\oplus$-attention by a large margin. Also, ELMo embeddings boost the performance.
This indicates a more challenging setup compared to TACRED, probably due to the higher number of \texttt{no-relation} instances and a large number of very similar evaluation instances that only differ in the choice of query entities.
Our model with enriched $\odot$-attention and ELMo embeddings sets the new state of the art for both relation extraction with gold and predicted entities.

\begin{figure}
	\begin{minipage}{.5\textwidth}
		\centering
		\includegraphics[width=.8\textwidth]{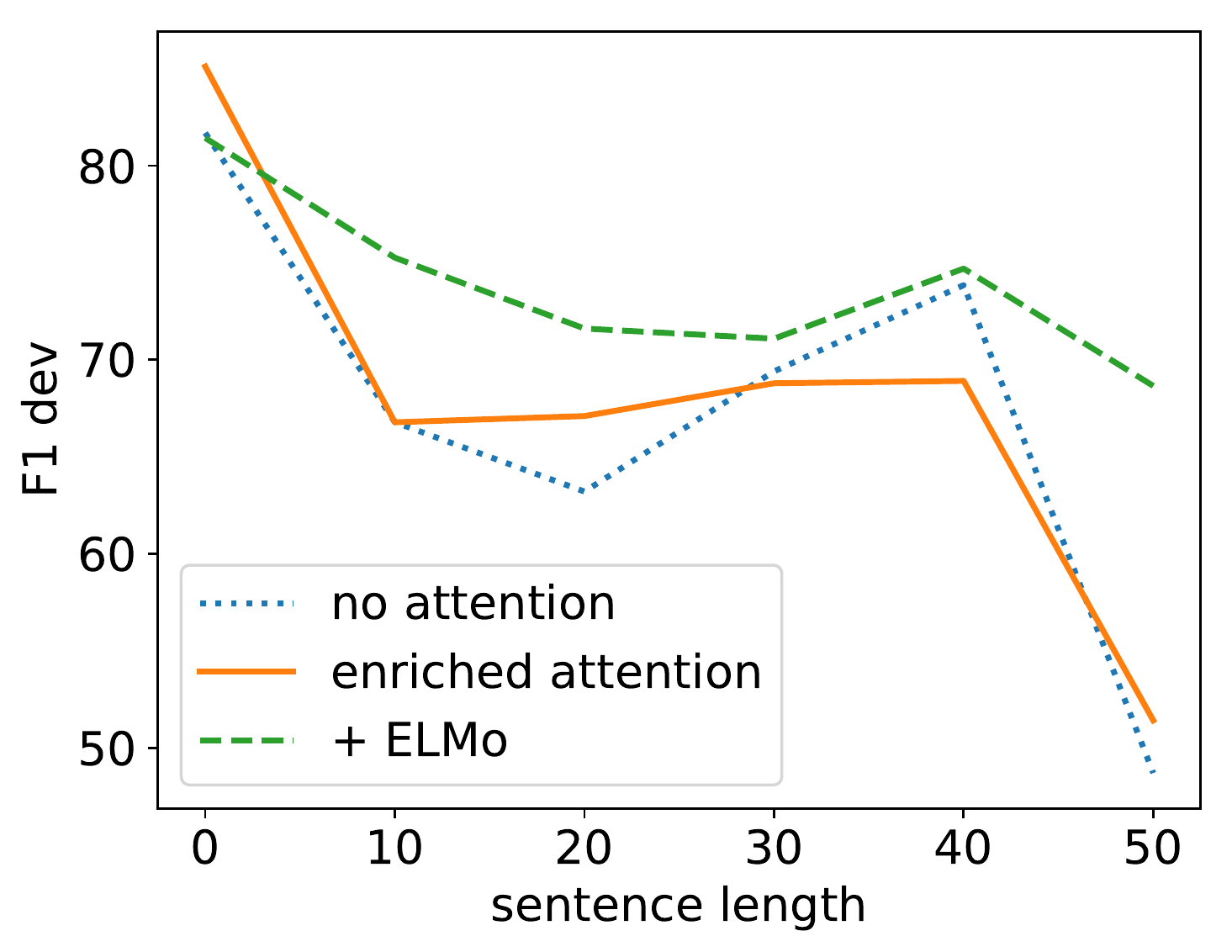}
		\caption{Performance with respect to sentence length on the ACE development set.}
		\label{fig:analysisACEquantitative1}
	\end{minipage}
	\begin{minipage}{.5\textwidth}
		\centering
		\includegraphics[width=.8\textwidth]{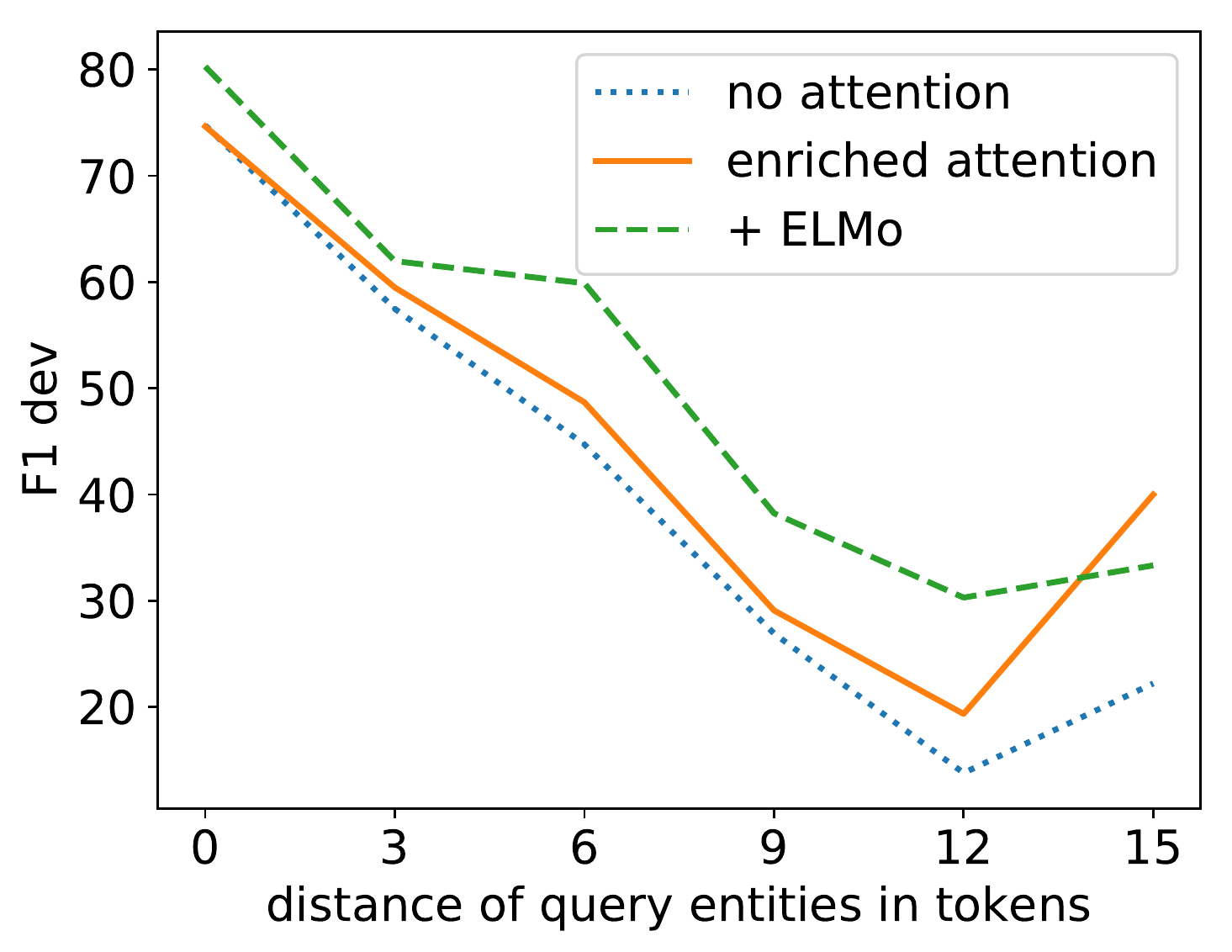}
		\caption{Performance with respect to distance of query entities on the ACE development set.}
		\label{fig:analysisACEquantitative2}
	\end{minipage}
	\begin{minipage}{.5\textwidth}
		\centering
		\includegraphics[width=.8\textwidth]{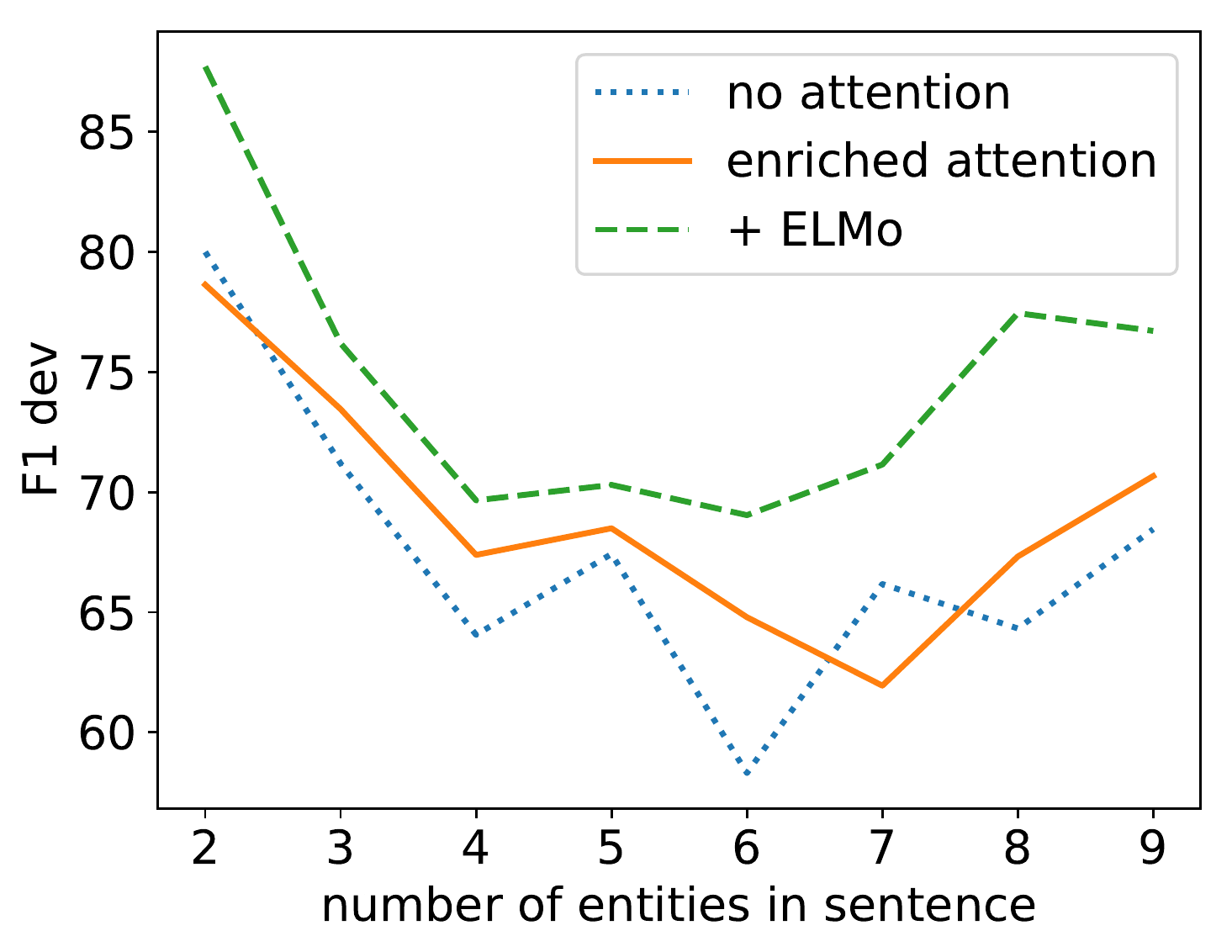}
		\caption{Performance with respect to number of entities in the sentence on the ACE development set.}
		\label{fig:analysisACEquantitative3}
	\end{minipage}
	\begin{minipage}{.5\textwidth}
		\centering
		\includegraphics[width=.8\textwidth]{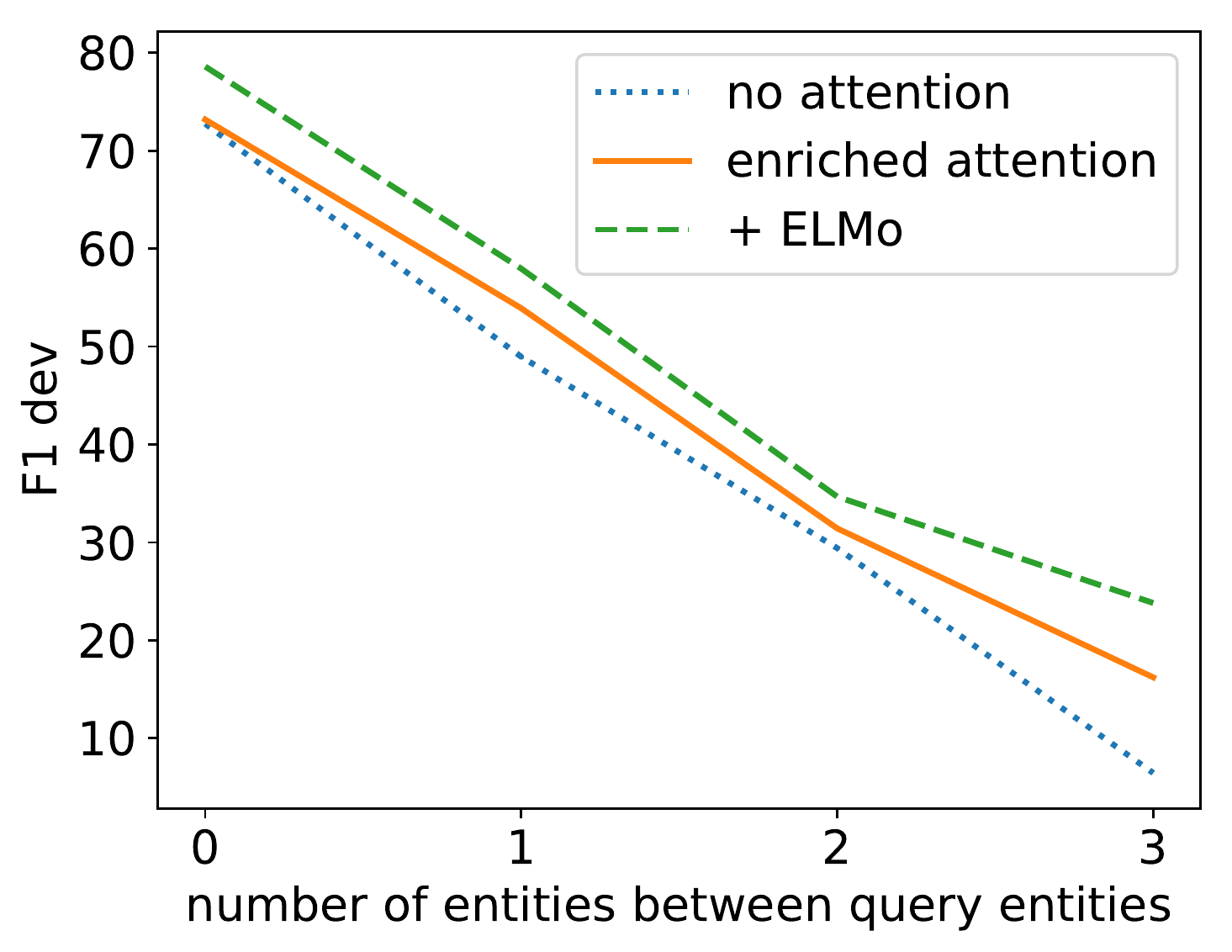}
		\caption{Performance with respect to number of entities between the query entities on the ACE development set.}
		\label{fig:analysisACEquantitative4}
	\end{minipage}
\end{figure}

\subsection{Analysis}
Next, we present a qualitative attention weight analysis and a quantitative robustness analysis.

\textbf{Attention weight analysis.}
Figure \ref{tab:analysis_weights} shows a heatmap of attention weights for the enriched attention model on a sample sentence of the ACE development set. The beginning and end of the sentence have attention weights of 0 and are cut for better readability.
We provide 
another
example in the appendix.
The different rows in the figure correspond to different query entities (marked as subject (``subj'') and object (``obj'')), the first column shows the gold relation label for the corresponding query entities which the model does not know but needs to predict.
The figure shows that depending on the query entities, the model attends to different parts of the input sentence.
It correctly focuses on those parts that are relevant for the query entities, such as ``of the Spanish media'' for the query entities ``members'' and ``media''
or ``and friends of the two members'' for the query entities ``family'' and ``members''.

\textbf{Robustness analysis.}
Figures \ref{fig:analysisACEquantitative1}-\ref{fig:analysisACEquantitative4} show the performance of our model with respect to sentence length, distance between query entities, number of entities per sentence and number of entities between the query entities on the ACE development set. For the former two, we count number of tokens and group the development set instances into bins (of size ten for sentence length and of size three for distance between query entities). To give an example, the F1 score shown for sentence length 10 corresponds to the micro F1 score of all dataset instances with length 10 to 19.
The graphs confirm that longer sentences with many entities
are more challenging, leading to lower $F_1$ scores.
Enriched attention
still suffers from this problem but increases performance, especially for sentences with many entities.

	\section{Conclusion}
In this paper, we proposed to enrich the attention layer of neural networks 
for relation extraction with additional features to increase their robustness and scalability in the case of long sentences with multiple entities. 
Our model outperforms prior work using comparable setups on the TACRED benchmark and achieves state-of-the-art results on the ACE 2005 benchmark.
Our analysis shows that it is indeed effective
for long sentences with many different entities, and correctly
focuses on the relevant parts of the sentence.
Directions for future work are the integration of enriched attention into transformer networks before pre-training as well as the exploration of enriched attention for other tasks.


\iftaclpubformat
\section*{Acknowledgments}
We would like to thank the members of the BCAI NLP\&KRR research group, especially Annemarie Friedrich, for their helpful comments.
\else
\fi

\bibliographystyle{acl_natbib}
\bibliography{refs}

\begin{thebibliography}{50}
\expandafter\ifx\csname natexlab\endcsname\relax\def\natexlab#1{#1}\fi

\bibitem[{Adel and Sch{\"u}tze(2017)}]{adel2017}
Heike Adel and Hinrich Sch{\"u}tze. 2017.
\newblock \href {https://www.aclweb.org/anthology/E17-1003} {Exploring
  different dimensions of attention for uncertainty detection}.
\newblock In \emph{Proceedings of the 15th Conference of the European Chapter
  of the Association for Computational Linguistics: Volume 1, Long Papers},
  pages 22--34, Valencia, Spain. Association for Computational Linguistics.

\bibitem[{Adel and Sch\"{u}tze(2019)}]{adel2019}
Heike Adel and Hinrich Sch\"{u}tze. 2019.
\newblock Type-aware convolutional neural networks for slot filling.
\newblock \emph{The Journal of Artificial Intelligence Research}, 66:297--339.

\bibitem[{Akbik et~al.(2018)Akbik, Blythe, and Vollgraf}]{akbik2018}
Alan Akbik, Duncan Blythe, and Roland Vollgraf. 2018.
\newblock \href {https://www.aclweb.org/anthology/C18-1139} {Contextual string
  embeddings for sequence labeling}.
\newblock In \emph{Proceedings of the 27th International Conference on
  Computational Linguistics}, pages 1638--1649, Santa Fe, New Mexico, USA.
  Association for Computational Linguistics.

\bibitem[{Alt et~al.(2019)Alt, H\"{u}bner, and Hennig}]{alt2019}
Christoph Alt, Marc H\"{u}bner, and Leonhard Hennig. 2019.
\newblock \href {https://openreview.net/forum?id=BJgrxbqp67} {Improving
  relation extraction by pre-trained language representations}.
\newblock In \emph{Proceedings of the 1st Conference on Automated Knowledge
  Base Construction (AKBC)}, Amherst, Massachusetts.

\bibitem[{Bahdanau et~al.(2015)Bahdanau, Cho, and Bengio}]{bahdanau2015}
Dzmitry Bahdanau, Kyunghyun Cho, and Yoshua Bengio. 2015.
\newblock \href {http://arxiv.org/abs/1409.0473} {Neural machine translation by
  jointly learning to align and translate}.
\newblock In \emph{3rd International Conference on Learning Representations,
  {ICLR} 2015}, San Diego, California.

\bibitem[{Bunescu and Mooney(2005)}]{bunescu2005}
Razvan~C. Bunescu and Raymond~J. Mooney. 2005.
\newblock A shortest path dependency kernel for relation extraction.
\newblock In \emph{Proceedings of the conference on Human Language}, pages
  724--731.

\bibitem[{Cai et~al.(2016)Cai, Zhang, and Wang}]{cai2016}
Rui Cai, Xiaodong Zhang, and Houfeng Wang. 2016.
\newblock \href {https://doi.org/10.18653/v1/P16-1072} {Bidirectional recurrent
  convolutional neural network for relation classification}.
\newblock In \emph{Proceedings of the 54th Annual Meeting of the Association
  for Computational Linguistics (Volume 1: Long Papers)}, pages 756--765,
  Berlin, Germany. Association for Computational Linguistics.

\bibitem[{Christopoulou et~al.(2018)Christopoulou, Miwa, and
  Ananiadou}]{christopoulou2019}
Fenia Christopoulou, Makoto Miwa, and Sophia Ananiadou. 2018.
\newblock \href {https://www.aclweb.org/anthology/P18-2014} {A walk-based model
  on entity graphs for relation extraction}.
\newblock In \emph{Proceedings of the 56th Annual Meeting of the Association
  for Computational Linguistics (Volume 2: Short Papers)}, pages 81--88,
  Melbourne, Australia. Association for Computational Linguistics.

\bibitem[{Conneau et~al.(2020)Conneau, Khandelwal, Goyal, Chaudhary, Wenzek,
  Guzm{\'a}n, Grave, Ott, Zettlemoyer, and Stoyanov}]{conneau2020}
Alexis Conneau, Kartikay Khandelwal, Naman Goyal, Vishrav Chaudhary, Guillaume
  Wenzek, Francisco Guzm{\'a}n, Edouard Grave, Myle Ott, Luke Zettlemoyer, and
  Veselin Stoyanov. 2020.
\newblock \href {https://www.aclweb.org/anthology/2020.acl-main.747}
  {Unsupervised cross-lingual representation learning at scale}.
\newblock In \emph{Proceedings of the 58th Annual Meeting of the Association
  for Computational Linguistics}, pages 8440--8451, Online. Association for
  Computational Linguistics.

\bibitem[{Del~Corro et~al.(2015)Del~Corro, Abujabal, Gemulla, and
  Weikum}]{DelCorroEtAl2015}
Luciano Del~Corro, Abdalghani Abujabal, Rainer Gemulla, and Gerhard Weikum.
  2015.
\newblock \href {https://doi.org/10.18653/v1/D15-1103} {{FINET}: Context-aware
  fine-grained named entity typing}.
\newblock In \emph{Proceedings of the 2015 Conference on Empirical Methods in
  Natural Language Processing}, pages 868--878, Lisbon, Portugal. Association
  for Computational Linguistics.

\bibitem[{Devlin et~al.(2019)Devlin, Chang, Lee, and Toutanova}]{devlin2019}
Jacob Devlin, Ming-Wei Chang, Kenton Lee, and Kristina Toutanova. 2019.
\newblock {BERT}: Pre-training of deep bidirectional transformers for language
  understanding.
\newblock In \emph{Proceedings of the 2019 Conference of the North {A}merican
  Chapter of the Association for Computational Linguistics: Human Language
  Technologies, Volume 1 (Long Papers)}, Minneapolis, Minnesota. Association
  for Computational Linguistics.

\bibitem[{Fundel et~al.(2006)Fundel, K{\"u}ffner, and Zimmer}]{fundel2006}
Katrin Fundel, Robert K{\"u}ffner, and Ralf Zimmer. 2006.
\newblock {RelEx} -- relation extraction using dependency parse trees.
\newblock \emph{Bioinformatics}, 23(3):365--371.

\bibitem[{Guo et~al.(2019)Guo, Zhang, and Lu}]{guo2019}
Zhijiang Guo, Yan Zhang, and Wei Lu. 2019.
\newblock \href {https://doi.org/10.18653/v1/P19-1024} {Attention guided graph
  convolutional networks for relation extraction}.
\newblock In \emph{Proceedings of the 57th Annual Meeting of the Association
  for Computational Linguistics}, pages 241--251, Florence, Italy. Association
  for Computational Linguistics.

\bibitem[{Heinzerling and Strube(2018)}]{heinzerling2018}
Benjamin Heinzerling and Michael Strube. 2018.
\newblock \href {https://www.aclweb.org/anthology/L18-1473} {{BPE}mb:
  Tokenization-free pre-trained subword embeddings in 275 languages}.
\newblock In \emph{Proceedings of the Eleventh International Conference on
  Language Resources and Evaluation ({LREC} 2018)}, Miyazaki, Japan. European
  Language Resources Association (ELRA).

\bibitem[{Hoffart et~al.(2011)Hoffart, Yosef, Bordino, F\"{u}rstenau, Pinkal,
  Spaniol, Taneva, Thater, and Weikum}]{HoffartEtAl2011}
Johannes Hoffart, Mohamed~Amir Yosef, Ilaria Bordino, Hagen F\"{u}rstenau,
  Manfred Pinkal, Marc Spaniol, Bilyana Taneva, Stefan Thater, and Gerhard
  Weikum. 2011.
\newblock \href {http://dl.acm.org/citation.cfm?id=2145432.2145521} {Robust
  disambiguation of named entities in text}.
\newblock In \emph{Proceedings of the Conference on Empirical Methods in
  Natural Language Processing}, EMNLP '11, pages 782--792, Stroudsburg, PA,
  USA. Association for Computational Linguistics.

\bibitem[{Huang et~al.(2017)Huang, Sil, Ji, and Florian}]{huang2017}
Lifu Huang, Avirup Sil, Heng Ji, and Radu Florian. 2017.
\newblock \href {https://doi.org/10.18653/v1/D17-1274} {Improving slot filling
  performance with attentive neural networks on dependency structures}.
\newblock In \emph{Proceedings of the 2017 Conference on Empirical Methods in
  Natural Language Processing}, pages 2588--2597, Copenhagen, Denmark.
  Association for Computational Linguistics.

\bibitem[{Lample et~al.(2016)Lample, Ballesteros, Subramanian, Kawakami, and
  Dyer}]{lample2016}
Guillaume Lample, Miguel Ballesteros, Sandeep Subramanian, Kazuya Kawakami, and
  Chris Dyer. 2016.
\newblock \href {https://doi.org/10.18653/v1/N16-1030} {Neural architectures
  for named entity recognition}.
\newblock In \emph{Proceedings of the 2016 Conference of the North {A}merican
  Chapter of the Association for Computational Linguistics: Human Language
  Technologies}, pages 260--270, San Diego, California. Association for
  Computational Linguistics.

\bibitem[{Li et~al.(2019{\natexlab{a}})Li, Mao, Yang, and Li}]{li2019}
Pengfei Li, Kezhi Mao, Xuefeng Yang, and Qi~Li. 2019{\natexlab{a}}.
\newblock \href {https://doi.org/10.18653/v1/D19-1022} {Improving relation
  extraction with knowledge-attention}.
\newblock In \emph{Proceedings of the 2019 Conference on Empirical Methods in
  Natural Language Processing and the 9th International Joint Conference on
  Natural Language Processing (EMNLP-IJCNLP)}, pages 229--239, Hong Kong,
  China. Association for Computational Linguistics.

\bibitem[{Li and Ji(2014)}]{li2014}
Qi~Li and Heng Ji. 2014.
\newblock \href {https://doi.org/10.3115/v1/P14-1038} {Incremental joint
  extraction of entity mentions and relations}.
\newblock In \emph{Proceedings of the 52nd Annual Meeting of the Association
  for Computational Linguistics (Volume 1: Long Papers)}, pages 402--412,
  Baltimore, Maryland. Association for Computational Linguistics.

\bibitem[{Li et~al.(2019{\natexlab{b}})Li, Yin, Sun, Li, Yuan, Chai, Zhou, and
  Li}]{li2019qa}
Xiaoya Li, Fan Yin, Zijun Sun, Xiayu Li, Arianna Yuan, Duo Chai, Mingxin Zhou,
  and Jiwei Li. 2019{\natexlab{b}}.
\newblock \href {https://doi.org/10.18653/v1/P19-1129} {Entity-relation
  extraction as multi-turn question answering}.
\newblock In \emph{Proceedings of the 57th Annual Meeting of the Association
  for Computational Linguistics}, pages 1340--1350, Florence, Italy.
  Association for Computational Linguistics.

\bibitem[{Lin et~al.(2020)Lin, Ji, Huang, and Wu}]{lin2020}
Ying Lin, Heng Ji, Fei Huang, and Lingfei Wu. 2020.
\newblock \href {https://www.aclweb.org/anthology/2020.acl-main.713} {A joint
  neural model for information extraction with global features}.
\newblock In \emph{Proceedings of the 58th Annual Meeting of the Association
  for Computational Linguistics}, pages 7999--8009, Online. Association for
  Computational Linguistics.

\bibitem[{Ling and Weld(2012)}]{ling2012}
Xiao Ling and Daniel~S. Weld. 2012.
\newblock Fine-grained entity recognition.
\newblock In \emph{Proceedings of the Twenty-Sixth AAAI Conference on
  Artificial Intelligence}, pages 94--100, Toronto, Ontario, Canada. AAAI
  Press.

\bibitem[{Luan et~al.(2019)Luan, Wadden, He, Shah, Ostendorf, and
  Hajishirzi}]{luan2019}
Yi~Luan, Dave Wadden, Luheng He, Amy Shah, Mari Ostendorf, and Hannaneh
  Hajishirzi. 2019.
\newblock \href {https://doi.org/10.18653/v1/N19-1308} {A general framework for
  information extraction using dynamic span graphs}.
\newblock In \emph{Proceedings of the 2019 Conference of the North {A}merican
  Chapter of the Association for Computational Linguistics: Human Language
  Technologies, Volume 1 (Long and Short Papers)}, pages 3036--3046,
  Minneapolis, Minnesota. Association for Computational Linguistics.

\bibitem[{Manning et~al.(2014)Manning, Surdeanu, Bauer, Finkel, Bethard, and
  McClosky}]{manning2014}
Christopher Manning, Mihai Surdeanu, John Bauer, Jenny Finkel, Steven Bethard,
  and David McClosky. 2014.
\newblock \href {https://doi.org/10.3115/v1/P14-5010} {The {S}tanford
  {C}ore{NLP} natural language processing toolkit}.
\newblock In \emph{Proceedings of 52nd Annual Meeting of the Association for
  Computational Linguistics: System Demonstrations}, pages 55--60, Baltimore,
  Maryland. Association for Computational Linguistics.

\bibitem[{Mintz et~al.(2009)Mintz, Bills, Snow, and Jurafsky}]{mintz2009}
Mike Mintz, Steven Bills, Rion Snow, and Daniel Jurafsky. 2009.
\newblock \href {https://www.aclweb.org/anthology/P09-1113} {Distant
  supervision for relation extraction without labeled data}.
\newblock In \emph{Proceedings of the Joint Conference of the 47th Annual
  Meeting of the {ACL} and the 4th International Joint Conference on Natural
  Language Processing of the {AFNLP}}, pages 1003--1011, Suntec, Singapore.
  Association for Computational Linguistics.

\bibitem[{Miwa and Bansal(2016)}]{miwa2016}
Makoto Miwa and Mohit Bansal. 2016.
\newblock \href {https://doi.org/10.18653/v1/P16-1105} {End-to-end relation
  extraction using {LSTMs} on sequences and tree structures}.
\newblock In \emph{Proceedings of the 54th Annual Meeting of the Association
  for Computational Linguistics (Volume 1: Long Papers)}, pages 1105--1116,
  Berlin, Germany. Association for Computational Linguistics.

\bibitem[{Miwa and Sasaki(2014)}]{miwa2014}
Makoto Miwa and Yutaka Sasaki. 2014.
\newblock \href {https://doi.org/10.3115/v1/D14-1200} {Modeling joint entity
  and relation extraction with table representation}.
\newblock In \emph{Proceedings of the 2014 Conference on Empirical Methods in
  Natural Language Processing ({EMNLP})}, pages 1858--1869, Doha, Qatar.
  Association for Computational Linguistics.

\bibitem[{Nguyen and Grishman(2016)}]{nguyen2016}
Thien~Huu Nguyen and Ralph Grishman. 2016.
\newblock Combining neural networks and log-linear models to improve relation
  extraction.
\newblock In \emph{Proceedings of the IJCAI Workshop on Deep Learning for
  Artificial Intelligence}.

\bibitem[{Pennington et~al.(2014)Pennington, Socher, and
  Manning}]{pennington2014}
Jeffrey Pennington, Richard Socher, and Christopher Manning. 2014.
\newblock \href {https://doi.org/10.3115/v1/D14-1162} {{GloVe}: Global vectors
  for word representation}.
\newblock In \emph{Proceedings of the 2014 Conference on Empirical Methods in
  Natural Language Processing ({EMNLP})}, pages 1532--1543, Doha, Qatar.
  Association for Computational Linguistics.

\bibitem[{Peters et~al.(2018)Peters, Neumann, Iyyer, Gardner, Clark, Lee, and
  Zettlemoyer}]{peters2018}
Matthew Peters, Mark Neumann, Mohit Iyyer, Matt Gardner, Christopher Clark,
  Kenton Lee, and Luke Zettlemoyer. 2018.
\newblock \href {https://doi.org/10.18653/v1/N18-1202} {Deep contextualized
  word representations}.
\newblock In \emph{Proceedings of the 2018 Conference of the North {A}merican
  Chapter of the Association for Computational Linguistics: Human Language
  Technologies, Volume 1 (Long Papers)}, pages 2227--2237, New Orleans,
  Louisiana. Association for Computational Linguistics.

\bibitem[{Ren et~al.(2017)Ren, Wu, He, Qu, Voss, Ji, Abdelzaher, and
  Han}]{ren2017}
Xiang Ren, Zeqiu Wu, Wenqi He, Meng Qu, Clare~R. Voss, Heng Ji, Tarek~F.
  Abdelzaher, and Jiawei Han. 2017.
\newblock Cotype: Joint extraction of typed entities and relations with
  knowledge bases.
\newblock In \emph{Proceedings of the 26th World Wide Web Conference}, Perth,
  Australia.

\bibitem[{Rink and Harabagiu(2010)}]{rink2010}
Bryan Rink and Sanda Harabagiu. 2010.
\newblock \href {https://www.aclweb.org/anthology/S10-1057} {{UTD}: Classifying
  semantic relations by combining lexical and semantic resources}.
\newblock In \emph{Proceedings of the 5th International Workshop on Semantic
  Evaluation}, pages 256--259, Uppsala, Sweden. Association for Computational
  Linguistics.

\bibitem[{Roth and Yih(2004)}]{roth2004}
Dan Roth and Wen-tau Yih. 2004.
\newblock \href {https://www.aclweb.org/anthology/W04-2401} {A linear
  programming formulation for global inference in natural language tasks}.
\newblock In \emph{Proceedings of the Eighth Conference on Computational
  Natural Language Learning ({C}o{NLL}-2004) at {HLT}-{NAACL} 2004}, pages
  1--8, Boston, Massachusetts, USA. Association for Computational Linguistics.

\bibitem[{Soares et~al.(2019)Soares, FitzGerald, Ling, and
  Kwiatkowski}]{soares2019}
Baldini~Livio Soares, Nicholas FitzGerald, Jeffrey Ling, and Tom Kwiatkowski.
  2019.
\newblock \href {https://doi.org/10.18653/v1/P19-1279} {Matching the blanks:
  Distributional similarity for relation learning}.
\newblock In \emph{Proceedings of the 57th Annual Meeting of the Association
  for Computational Linguistics}, pages 2895--2905, Florence, Italy.
  Association for Computational Linguistics.

\bibitem[{Srivastava et~al.(2014)Srivastava, Hinton, Krizhevsky, Sutskever, and
  Salakhutdinov}]{srivastava2014}
Nitish Srivastava, Geoffrey Hinton, Alex Krizhevsky, Ilya Sutskever, and Ruslan
  Salakhutdinov. 2014.
\newblock \href {http://jmlr.org/papers/v15/srivastava14a.html} {Dropout: A
  simple way to prevent neural networks from overfitting}.
\newblock \emph{Journal of Machine Learning Research}, 15:1929--1958.

\bibitem[{Sun et~al.(2019)Sun, Gong, Wu, Gong, Jiang, Lan, Sun, and
  Duan}]{sun2019}
Changzhi Sun, Yeyun Gong, Yuanbin Wu, Ming Gong, Daxin Jiang, Man Lan, Shiliang
  Sun, and Nan Duan. 2019.
\newblock \href {https://doi.org/10.18653/v1/P19-1131} {Joint type inference on
  entities and relations via graph convolutional networks}.
\newblock In \emph{Proceedings of the 57th Annual Meeting of the Association
  for Computational Linguistics}, pages 1361--1370, Florence, Italy.
  Association for Computational Linguistics.

\bibitem[{Sun et~al.(2018)Sun, Wu, Lan, Sun, Wang, Lee, and Wu}]{sun2018}
Changzhi Sun, Yuanbin Wu, Man Lan, Shiliang Sun, Wenting Wang, Kuang-Chih Lee,
  and Kewen Wu. 2018.
\newblock \href {https://doi.org/10.18653/v1/D18-1249} {Extracting entities and
  relations with joint minimum risk training}.
\newblock In \emph{Proceedings of the 2018 Conference on Empirical Methods in
  Natural Language Processing}, pages 2256--2265, Brussels, Belgium.
  Association for Computational Linguistics.

\bibitem[{Toutanova et~al.(2015)Toutanova, Chen, Pantel, Poon, Choudhury, and
  Gamon}]{toutanova2015}
Kristina Toutanova, Danqi Chen, Patrick Pantel, Hoifung Poon, Pallavi
  Choudhury, and Michael Gamon. 2015.
\newblock \href {https://doi.org/10.18653/v1/D15-1174} {Representing text for
  joint embedding of text and knowledge bases}.
\newblock In \emph{Proceedings of the 2015 Conference on Empirical Methods in
  Natural Language Processing}, pages 1499--1509, Lisbon, Portugal. Association
  for Computational Linguistics.

\bibitem[{Vaswani et~al.(2017)Vaswani, Shazeer, Parmar, Uszkoreit, Jones,
  Gomez, Kaiser, and Polosukhin}]{vaswani2017}
Ashish Vaswani, Noam Shazeer, Niki Parmar, Jakob Uszkoreit, Llion Jones,
  Aidan~N Gomez, {\L}ukasz Kaiser, and Illia Polosukhin. 2017.
\newblock \href
  {http://papers.nips.cc/paper/7181-attention-is-all-you-need.pdf} {Attention
  is all you need}.
\newblock In I.~Guyon, U.~V. Luxburg, S.~Bengio, H.~Wallach, R.~Fergus,
  S.~Vishwanathan, and R.~Garnett, editors, \emph{Advances in Neural
  Information Processing Systems 30}, pages 5998--6008. Curran Associates, Inc.

\bibitem[{Veyseh et~al.(2020)Veyseh, Dernoncourt, Dou, and Nguyen}]{pouran2020}
Amir Pouran~Ben Veyseh, Franck Dernoncourt, Dejing Dou, and Thien~Huu Nguyen.
  2020.
\newblock \href {https://www.aclweb.org/anthology/2020.acl-main.715}
  {Exploiting the syntax-model consistency for neural relation extraction}.
\newblock In \emph{Proceedings of the 58th Annual Meeting of the Association
  for Computational Linguistics}, pages 8021--8032, Online. Association for
  Computational Linguistics.

\bibitem[{Walker et~al.(2006)Walker, Strassel, Medero, and Maeda}]{walker2006}
Christopher Walker, Stephanie Strassel, Julie Medero, and Kazuaki Maeda. 2006.
\newblock {ACE} 2005 multilingual training corpus.
\newblock \emph{Linguistic Data Consortium}, 57.

\bibitem[{Xu et~al.(2015)Xu, Mou, Li, Chen, Peng, and Jin}]{xu2015}
Yan Xu, Lili Mou, Ge~Li, Yunchuan Chen, Hao Peng, and Zhi Jin. 2015.
\newblock \href {https://doi.org/10.18653/v1/D15-1206} {Classifying relations
  via long short term memory networks along shortest dependency paths}.
\newblock In \emph{Proceedings of the 2015 Conference on Empirical Methods in
  Natural Language Processing}, pages 1785--1794, Lisbon, Portugal. Association
  for Computational Linguistics.

\bibitem[{Yaghoobzadeh et~al.(2018)Yaghoobzadeh, Adel, and
  Sch\"{u}tze}]{yaghoobzadeh2018}
Yadollah Yaghoobzadeh, Heike Adel, and Hinrich Sch\"{u}tze. 2018.
\newblock Corpus-level fine-grained entity typing.
\newblock \emph{Journal of Artificial Intelligence Research}, 61:835--862.

\bibitem[{Yamada et~al.(2017)Yamada, Shindo, Takeda, and Takefuji}]{yamada2017}
Ikuya Yamada, Hiroyuki Shindo, Hideaki Takeda, and Yoshiyasu Takefuji. 2017.
\newblock \href {https://doi.org/10.1162/tacl_a_00069} {Learning distributed
  representations of texts and entities from knowledge base}.
\newblock \emph{Transactions of the Association for Computational Linguistics},
  5:397--411.

\bibitem[{Yao et~al.(2010)Yao, Riedel, and McCallum}]{yao2010}
Limin Yao, Sebastian Riedel, and Andrew McCallum. 2010.
\newblock \href {https://www.aclweb.org/anthology/D10-1099} {Collective
  cross-document relation extraction without labelled data}.
\newblock In \emph{Proceedings of the 2010 Conference on Empirical Methods in
  Natural Language Processing}, pages 1013--1023, Cambridge, Massachusetts.
  Association for Computational Linguistics.

\bibitem[{Ye et~al.(2019)Ye, Li, Xie, Sheng, Chen, and Zhang}]{ye2019}
Wei Ye, Bo~Li, Rui Xie, Zhonghao Sheng, Long Chen, and Shikun Zhang. 2019.
\newblock \href {https://doi.org/10.18653/v1/P19-1130} {Exploiting entity {BIO}
  tag embeddings and multi-task learning for relation extraction with
  imbalanced data}.
\newblock In \emph{Proceedings of the 57th Annual Meeting of the Association
  for Computational Linguistics}, pages 1351--1360, Florence, Italy.
  Association for Computational Linguistics.

\bibitem[{Zeng et~al.(2014)Zeng, Liu, Lai, Zhou, and Zhao}]{zeng2014}
Daojian Zeng, Kang Liu, Siwei Lai, Guangyou Zhou, and Jun Zhao. 2014.
\newblock \href {https://www.aclweb.org/anthology/C14-1220} {Relation
  classification via convolutional deep neural network}.
\newblock In \emph{Proceedings of COLING 2014, the 25th International
  Conference on Computational Linguistics: Technical Papers}, pages 2335--2344,
  Dublin, Ireland. Dublin City University and Association for Computational
  Linguistics.

\bibitem[{Zhang et~al.(2018)Zhang, Qi, and Manning}]{zhang2018}
Yuhao Zhang, Peng Qi, and Christopher~D. Manning. 2018.
\newblock \href {https://www.aclweb.org/anthology/D18-1244} {Graph convolution
  over pruned dependency trees improves relation extraction}.
\newblock In \emph{Proceedings of the 2018 Conference on Empirical Methods in
  Natural Language Processing}, pages 2205--2215, Brussels, Belgium.
  Association for Computational Linguistics.

\bibitem[{Zhang et~al.(2017)Zhang, Zhong, Chen, Angeli, and
  Manning}]{zhang2017}
Yuhao Zhang, Victor Zhong, Danqi Chen, Gabor Angeli, and Christopher~D.
  Manning. 2017.
\newblock \href {https://doi.org/10.18653/v1/D17-1004} {Position-aware
  attention and supervised data improve slot filling}.
\newblock In \emph{Proceedings of the 2017 Conference on Empirical Methods in
  Natural Language Processing}, pages 35--45, Copenhagen, Denmark. Association
  for Computational Linguistics.

\bibitem[{Zhong et~al.(2019)Zhong, Wang, and Miao}]{zhong2019}
Peixiang Zhong, Di~Wang, and Chunyan Miao. 2019.
\newblock \href {https://doi.org/10.18653/v1/D19-1016} {Knowledge-enriched
  transformer for emotion detection in textual conversations}.
\newblock In \emph{Proceedings of the 2019 Conference on Empirical Methods in
  Natural Language Processing and the 9th International Joint Conference on
  Natural Language Processing (EMNLP-IJCNLP)}, pages 165--176, Hong Kong,
  China. Association for Computational Linguistics.

\end{thebibliography}

\section{Appendix}

\subsection{Type Mapping}
To represent out-of-knowledge base entities with Wikipedia embeddings, we
utilize their type as described in Section 4.2 of the main paper.
Table \ref{tab:mapping} provides the mapping of types to Wikipedia pages.

\begin{table}[h]
\footnotesize
\centering
 \begin{tabular}{p{.01cm}l|l}
  & Type & Wikipedia page\\
  \hline 
  \multirow{7}{*}{\rotatebox{90}{ACE 2005}} &
  PER & Person\\
  &ORG & Organization\\
  &LOC & Location (geography)\\
  &GPE & Nation\\
  &FAC & Physical plant\\
  &VEH & Vehicle\\
  &WEA & Weapon\\
  \hline 
  \multirow{17}{*}{\rotatebox{90}{TACRED}} &
  PERSON & Person\\
  &ORGANIZATION & Organization\\
  &LOCATION & Location (geography)\\
  &CITY & City \\
  &STATE\_OR\_PROVINCE & State (polity)\\
  &COUNTRY & Country \\
  &CAUSE\_OF\_DEATH & Cause of death\\
  &CRIMINAL\_CHARGE & Criminal charge\\
  &DATE & Calendar date\\
  &DURATION & Time\\
  &IDEOLOGY & Ideology\\
  &NATIONALITY & Nationality \\
  &NUMBER & Number \\
  &RELIGION & Religion\\
  &TITLE & Profession\\
  &URL & Uniform Resource Locator
 \end{tabular}
\caption{Mapping of types (for datasets ACE 2005 and TACRED) to Wikipedia pages.}
\label{tab:mapping}
\end{table}

\begin{figure*}
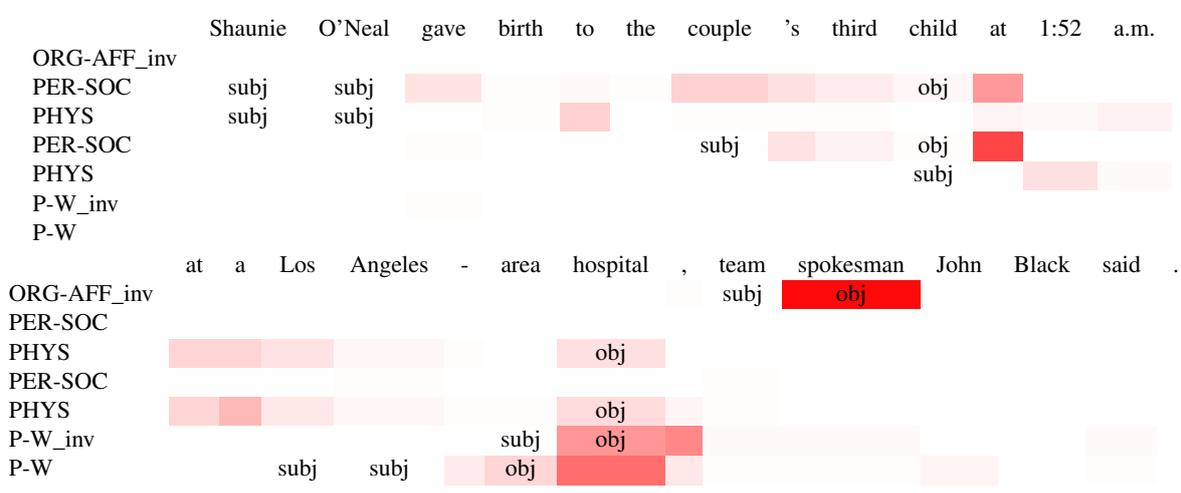

	\footnotesize
	\centering
	\begin{tabular}{lccccccccccccc}
		& Shaunie & O'Neal & gave & birth & to & the & couple & 's & third & child & at & 1:52 & a.m.\\
		ORG-AFF\_inv & \cellcolor{red!0} & \cellcolor{red!0} & \cellcolor{red!0} & \cellcolor{red!0} & \cellcolor{red!0} & \cellcolor{red!0} & \cellcolor{red!0} & \cellcolor{red!0} & \cellcolor{red!0} & \cellcolor{red!0} & \cellcolor{red!0} & \cellcolor{red!0} & \cellcolor{red!0} \\
		PER-SOC & \cellcolor{red!0}subj & \cellcolor{red!0}subj & \cellcolor{red!11} & \cellcolor{red!1} & \cellcolor{red!2} & \cellcolor{red!1} & \cellcolor{red!18} & \cellcolor{red!12} &\cellcolor{red!7} & \cellcolor{red!3}obj & \cellcolor{red!40} & \cellcolor{red!0} & \cellcolor{red!0}\\
		PHYS & \cellcolor{red!0}subj & \cellcolor{red!0}subj & \cellcolor{red!0} & \cellcolor{red!1} & \cellcolor{red!18} & \cellcolor{red!0}  & \cellcolor{red!1} & \cellcolor{red!1} & \cellcolor{red!1} & \cellcolor{red!0} & \cellcolor{red!4} & \cellcolor{red!2} & \cellcolor{red!5}\\
		PER-SOC & \cellcolor{red!0} & \cellcolor{red!0} & \cellcolor{red!1} & \cellcolor{red!0} & \cellcolor{red!0} & \cellcolor{red!0} & \cellcolor{red!0}subj & \cellcolor{red!11} & \cellcolor{red!5} & \cellcolor{red!1}obj & \cellcolor{red!73} & \cellcolor{red!0} & \cellcolor{red!0}\\
		PHYS & \cellcolor{red!0} & \cellcolor{red!0} & \cellcolor{red!0} & \cellcolor{red!0} & \cellcolor{red!0} & \cellcolor{red!0}& \cellcolor{red!0} & \cellcolor{red!0} & \cellcolor{red!0} & \cellcolor{red!0}subj & \cellcolor{red!0} & \cellcolor{red!12} & \cellcolor{red!2}\\
		P-W\_inv & \cellcolor{red!0} & \cellcolor{red!0} & \cellcolor{red!1} & \cellcolor{red!0} & \cellcolor{red!0} & \cellcolor{red!0} & \cellcolor{red!0} & \cellcolor{red!0}{ } & \cellcolor{red!0} & \cellcolor{red!0} & \cellcolor{red!0} & \cellcolor{red!0} & \cellcolor{red!0}\\
		P-W & \cellcolor{red!0} & \cellcolor{red!0} & \cellcolor{red!0} & \cellcolor{red!0} & \cellcolor{red!0} & \cellcolor{red!0}  & \cellcolor{red!0}  &\cellcolor{red!0} & \cellcolor{red!0} & \cellcolor{red!0} & \cellcolor{red!0} & \cellcolor{red!0} & \cellcolor{red!0}
	\end{tabular}
	
	\begin{tabular}{lcccccccccccccc}
		& at & a & Los & Angeles & - & area & hospital & , & team & spokesman & John & Black & said & .\\
		ORG-AFF\_inv &  \cellcolor{red!0} & \cellcolor{red!0}{ }& \cellcolor{red!0} & \cellcolor{red!0} & \cellcolor{red!0} & \cellcolor{red!0} & \cellcolor{red!0} & \cellcolor{red!1} & \cellcolor{red!0}subj & \cellcolor{red!96}obj & \cellcolor{red!0} & \cellcolor{red!0} & \cellcolor{red!0} & \cellcolor{red!0}\\
		PER-SOC  & \cellcolor{red!0} & \cellcolor{red!0}{ } & \cellcolor{red!0} & \cellcolor{red!0} & \cellcolor{red!0} & \cellcolor{red!0} & \cellcolor{red!0} & \cellcolor{red!0} & \cellcolor{red!0} & \cellcolor{red!0} & \cellcolor{red!0} & \cellcolor{red!0} & \cellcolor{red!0} & \cellcolor{red!0}\\
		PHYS  & \cellcolor{red!17} & \cellcolor{red!17} & \cellcolor{red!11} & \cellcolor{red!3} & \cellcolor{red!1} & \cellcolor{red!0} & \cellcolor{red!12}obj & \cellcolor{red!0} & \cellcolor{red!0} & \cellcolor{red!0} & \cellcolor{red!0} & \cellcolor{red!0} & \cellcolor{red!0} & \cellcolor{red!0}\\
		PER-SOC  & \cellcolor{red!0}  & \cellcolor{red!0} \cellcolor{red!0} & \cellcolor{red!0} & \cellcolor{red!1} & \cellcolor{red!0} & \cellcolor{red!0} & \cellcolor{red!0} & \cellcolor{red!0} & \cellcolor{red!1} & \cellcolor{red!0} & \cellcolor{red!0} & \cellcolor{red!0} & \cellcolor{red!0}\\
		PHYS  & \cellcolor{red!16} & \cellcolor{red!28} & \cellcolor{red!9} & \cellcolor{red!3} & \cellcolor{red!1} & \cellcolor{red!1} & \cellcolor{red!14}obj & \cellcolor{red!4} & \cellcolor{red!1} & \cellcolor{red!0} & \cellcolor{red!0} & \cellcolor{red!0} & \cellcolor{red!0} & \cellcolor{red!0}\\
		P-W\_inv  & \cellcolor{red!0} & \cellcolor{red!0} & \cellcolor{red!0} & \cellcolor{red!0} & \cellcolor{red!0} & \cellcolor{red!0}subj & \cellcolor{red!41}obj & \cellcolor{red!47} & \cellcolor{red!2} & \cellcolor{red!2} & \cellcolor{red!0} & \cellcolor{red!0} & \cellcolor{red!2} & \cellcolor{red!0}\\
		P-W  & \cellcolor{red!0} & \cellcolor{red!0} & \cellcolor{red!0}subj & \cellcolor{red!0}subj & \cellcolor{red!8} & \cellcolor{red!16}obj & \cellcolor{red!57} & \cellcolor{red!9} & \cellcolor{red!1} & \cellcolor{red!1} & \cellcolor{red!4} & \cellcolor{red!0} & \cellcolor{red!1} & \cellcolor{red!0}
	\end{tabular}
	\caption{Attention weights for example sentence with different query entities from ACE development set. First column shows gold relation labels (not known to the model). P-W: PART-WHOLE}
	\label{tab:analysis_weights2}
\end{figure*}

\subsection{Attention Weight Analysis}

Figure \ref{tab:analysis_weights2} shows the same analyses as Figure 3 in the main paper for another sentence containing even more gold relations.

\end{document}